\documentclass[conference]{IEEEtran}
\IEEEoverridecommandlockouts
\usepackage[hidelinks]{hyperref}
\usepackage{cite}
\usepackage{amsmath,amssymb,amsfonts}
\usepackage{algorithmic}
\usepackage{graphicx}
\usepackage{textcomp}
\usepackage{subcaption}
\usepackage{xcolor}
\usepackage{orcidlink}

\def\BibTeX{{\rm B\kern-.05em{\sc i\kern-.025em b}\kern-.08em
    T\kern-.1667em\lower.7ex\hbox{E}\kern-.125emX}}

\newcommand{\SMART}{S.M.A.R.T. }
    
\begin{document}

\title{TFBEST: Dual-Aspect Transformer with Learnable Positional Encoding for Failure Prediction}

\author{\IEEEauthorblockN{Rohan Mohapatra \orcidlink{0000-0003-1654-7994}\IEEEauthorrefmark{1} and
Saptarshi Sengupta \orcidlink{0000-0003-1114-343X}\IEEEauthorrefmark{1}}
\IEEEauthorblockA{\IEEEauthorrefmark{1}Department of Computer Science, San Jos\'e State University, San Jos\'e, CA, USA \\
Email: \IEEEauthorrefmark{1}rohan.mohapatra@sjsu.edu,
\IEEEauthorrefmark{1}saptarshi.sengupta@sjsu.edu}}

\maketitle

\begin{abstract}
Hard Disk Drive (HDD) failures in datacenters are costly - from catastrophic data loss to a question of goodwill, stakeholders want to avoid it like the plague. An important tool in proactively monitoring against HDD failure is timely estimation of the Remaining Useful Life (RUL). To this end, the Self-Monitoring, Analysis and Reporting Technology employed within HDDs (S.M.A.R.T.) provide critical logs for long-term maintenance of the security and dependability of these essential data storage devices. Data-driven predictive models in the past have used these S.M.A.R.T. logs and CNN/RNN based architectures heavily. However, they have suffered significantly in providing a confidence interval around the predicted RUL values as well as in processing very long sequences of logs. In addition, some of these approaches, such as those based on LSTMs, are inherently slow to train and have tedious feature engineering overheads. To overcome these challenges, in this work we propose a novel transformer architecture - a Temporal-fusion Bi-encoder Self-attention Transformer (TFBEST) for predicting failures in hard-drives. It is an encoder-decoder based deep learning technique that enhances the context gained from understanding health statistics sequences and predicts a sequence of the number of days remaining before a disk potentially fails. In this paper, we also provide a novel confidence margin statistic that can help manufacturers replace a hard-drive within a time frame. Experiments on Seagate HDD data show that our method significantly outperforms the state-of-the-art RUL prediction methods during testing over the exhaustive 10-year data from Backblaze (2013-present). Although validated on HDD failure prediction, the TFBEST architecture is well-suited for other prognostics applications and may be adapted for allied regression problems.
\end{abstract}


\section{Introduction}
Condition-based maintenance (CBM) of equipment in large-scale cyber-physical systems (CPS) aims to maximize the reliability of these systems by selectively replacing parts that are predicted to fail\cite{BASAK2021101283}. The underlying assumption is that the degradation model these systems follow show up as trends in data that can be effectively used to predict the Remaining Useful Life (RUL) of such systems. Hard disk drives (HDD) are an important storage component of many systems, from personal computers to distributed data-centers. An HDD failure in a data-center can lead to catastrophic data loss if pertinent backup plans are not maintained. It is important to keep in mind that electro-mechanical devices are prone to failure owing to varied operational conditions and aging, therefore HDDs are no exception. However, early sensing of triggers and out-of-distribution signatures in the data can greatly help plan for failure, in the event it occurs. One common measure used for fault quantification in such cases is the Annualized Failure Rate (AFR). AFR represents the likelihood, expressed as a percentage, of a drive experiencing failure within a year. This probability is derived from the performance patterns of comparable drives. 

The informed reader might be aware that  prognostication techniques, as of this date, are broadly divided into three categories: model-based, data-driven and hybrid. Model-based techniques require precise and thorough dynamic modeling of electro-mechanical machinery or of individual components therein. This becomes an extremely difficult task if there is non-linearity in the system dynamics. Data-driven approaches, on the other hand, exploit the wealth of sensor data features to understand complex interrelationships between these features and use it to estimate the likelihood of failure in future. In the case of HDDs, data-driven approaches look at  \SMART sensor features to model the dependencies and look for patterns of interest. Of course, one might look at hybrid approaches as well, where the Physics of Failure is captured by a joint modeling using dynamical system equations and machine learning models \cite{physics}.

The \SMART monitoring system is essential for protecting HDDs over the course of their useful lives. Temperature, operation hours, on/off cycles, damaged sectors, and read/write errors are just a few of the many variables that it meticulously tracks. Continuous comparisons are made between these metrics and predetermined thresholds established by the HDD manufacturers. The system warns the user proactively when a particular \SMART parameter exceeds its set threshold. A laudable safety measure, this prompt notification enables users to take preventive actions like data backup and prompt drive replacement, preventing potential data loss. But relying on thresholds can lead to very premature drive change.  In the literature, machine learning algorithms have been used for RUL prediction, such as Support Vector Machines (SVM) \cite{svm}, Hidden Markov-models (HMM) \cite{hmmhdd} and Long short term memory networks (LSTM)\cite{lstm-sengupta}. These techniques, however, rely on time-consuming feature engineering. Transformer \cite{vaswani_trans} based approaches, in comparison, can automatically extract useful features from the data and achieve significantly superior prediction performance. 

Another notable issue is that more attention should be paid to the essential features providing significant information on degradation. Transformer architectures \cite{vaswani_trans} using the attention mechanism \cite{bahdanau2016neural} can learn the degradation mechanics without feature selection and can tend to long input sequences. Very recently, the Dual Aspect Self-Attention Transformer (DAST) \cite{dast} has been proposed, which adds an additional encoder over the vanilla transformer \cite{vaswani_trans} to extract more information about how different sensors affect the system rather than attending to only the weights of different time steps. However, the DAST network uses an absolute positional encoding which can not effectively model the timesteps. To overcome this, we propose and extensively validate a novel transformer architecture: Temporal-Fusion Bi-Encoder Self-attention Transformer (TFBEST) that uses a learnable positional encoding to encode position of the timesteps in the encoder.

\textbf{Contributions:} The contributions of this work are significant in the following ways:

\begin{enumerate}
    \item \textit{\textbf{A novel transformer architecture for remaining useful life prediction:}} We propose a new Transformer architecture which employs a learnable positional encoding to encode position of the timesteps in the encoder and two encoders (sensor and time) to extract information in parallel which avoids mutual influence of information from two aspects. To the best of our knowledge, transformers has not been implemented before for HDD RUL prediction using the comprehensive 10-year quarter-by-quarter health data from Backblaze \cite{backblazeHardDrive}. We show that our new network outperforms state-of-the-art architectures and produces highly accurate point estimates of the RUL.
    \item \textit{\textbf{Attention Mechanism instead of Feature Engineering:}} Our contribution also lies in using an attention mechanism for the encoder-decoder netowrk, This removes the need of feature selection as explored in \cite{rohan_1}.
    \item \textit{\textbf{Confidence interval and error margin evaluation:}} We propose a new confidence interval evaluation to provide a robust range of RUL for a hard-drive. This instils confidence around when a hard-drive will fail.

\end{enumerate}

The paper is organized in the following way. In Section \ref{section:lit}, we go over related work in the field. Section \ref{section:attention} provides an overview of why feature engineering is tedious and the shift to attention mechanisms. Section \ref{section:trans} talks about the details and architecture of the Vanilla Transformer. Section \ref{section:tfbest} introduces the proposed method and Section \ref{section:experiments} divulges into the different configurations and experimentation done with the dataset. Lastly, Section \ref{section:future} looks at potential avenues for future research.

\section{Related Work}
\label{section:lit}

Self-Monitoring, Analysis and Reporting Technology (\SMART) is a monitoring system included in hard disk drives (HDDs). Primarily, it senses and creates logs of various health indicators of drives thereby enabling proactive planning against imminent hardware failures.

Many data-driven approaches use \SMART features as their input data and provide predictions on the RUL of the hard-drives. The architectures based on deep learning can automatically collect the key feature information from the original data to perform end-to-end prediction by modeling the functional relationship between the hard-drive degradation process and the \SMART monitoring data. 

To date, many deep-learning architectures have been proposed for RUL predictions: TCNN \cite{luetal}, LSTM \cite{luetal}, Stacked LSTM \cite{stackedlstm}, Bi-directional LSTM \cite{coursey_lstm}, Spatio-temporal Anomaly Detection LSTM Networks \cite{BASAK2021101283}, Encoder-Decoder LSTM \cite{rohan_1}, and Ensemble Learning approaches \cite{ensemble}. All of these architectures have modeled the dataset in a specific way and have applied feature engineering to sanitize the data. However, the literature does not contain a case where attention mechanisms and transformers have been used for remaining useful life (RUL) prediction of hard disk drives. This opens up an avenue to apply transformers for prognosticating impending faults in HDDs and reporting the same.

Transformers are appealing to time series problems such as RUL estimation as they have demonstrated excellent modeling capability for long-range dependencies and interactions in sequential data. The vanilla transformer has undergone many iterations that have been used for a variety of time series activities to meet unique issues in forecasting. 

In recent studies, there have been examples of different variants of Transformers employed for specific tasks like forecasting \cite{lietal, fedformer} and anomaly detection \cite{xu2022anomaly, tuli}. Specifically, seasonality or periodicity is an important feature of time series observations \cite{wenetal} leveraged by transformers for generating predictions\cite{dast}.

\section{Caveats of feature engineering \& Shift towards attention mechanism}
\label{section:attention}
\subsection{Problems with feature selection}
Feature engineering plays a fundamental role in the process of getting data ready for machine learning and data analysis. It entails generating novel features or transforming existing ones to boost the performance of a machine learning model, or allowing greater insight from the data. In our prior work \cite{rohan_1}, we used a subset of features for the same task in order to make things simpler and to filter out redundant attributes. But feature selection is a tedious process: (a) It requires understanding data and its trends. (b) It can be somewhat unreliable, since minute alterations to either the dataset or how the features are chosen can lead to a completely different group of features being chosen. (c) It also becomes extremely challenging to be aware of the consequences of maximizing information gain during feature selection, such as potential overfitting particularly when dealing with noisy data such as the one from Backblaze. We must take into account the trade-off between reducing bias and increasing variance for selecting features in order to make an informed decision. 

\subsection{Attention Mechanism can replace feature selection}
Transformers have revolutionized natural language processing (NLP) \cite{transformers_survey} and have found applications in other research areas. Because of their ability to automatically learn useful properties from data, Transformers' attention mechanism, particularly in models such as Bidirectional Encoder Representations from Transformers (BERT) \cite{bert} and Generative Pre-trained Transformer (GPT) \cite{gpt1}, enable them to capture intricate correlations in data \cite{lin2022feature}.

\section{Preliminaries of the Transformer}
\label{section:trans}
\subsection{The Vanilla Transformer}
\begin{figure}[htbp]
\centering
\includegraphics[width=0.7\columnwidth]{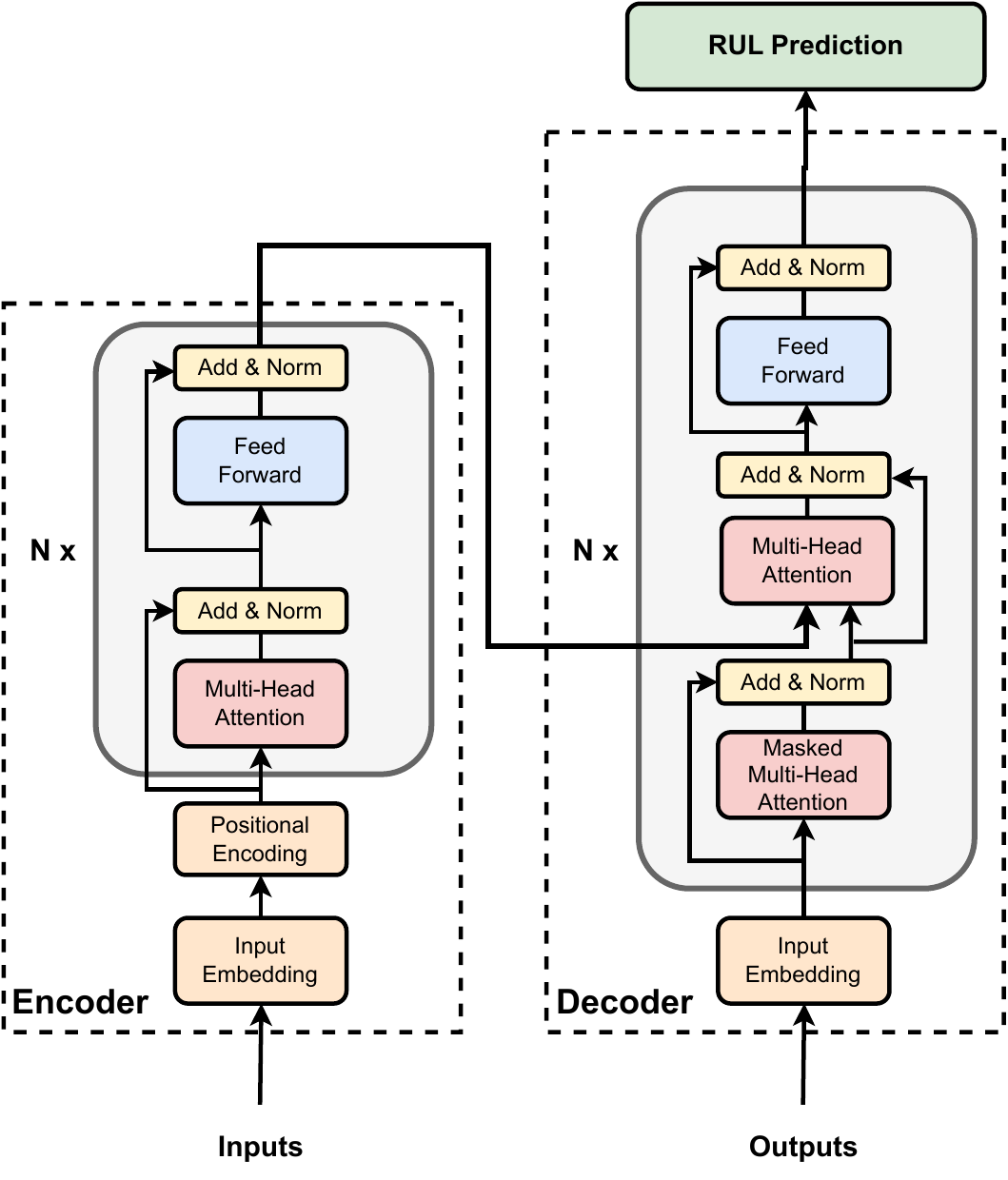}
\caption{Architecture of Vanilla Transformer }
\label{fig:vanillatransformer}
\end{figure}

The vanilla Transformer \cite{vaswani_trans} was the first to introduce the attention mechanism. It has been very successful in NLP tasks. As shown in the Fig. \ref{fig:vanillatransformer}, it follows an encoder-decoder structure where both the encoder and the decoder are made up of numerous identical blocks. Each encoder block is made up of a multi-head self-attention module and a position-wise feed-forward network, while each decoder block is made up of cross-attention models that are inserted between the multi-head self-attention module and the position-wise feed-forward network. 

The vanilla Transformer, unlike an LSTM or RNN, has no recurrence. Instead, it models the sequence information using the positional encoding included in the input embeddings. This can be beneficial for NLP and Neural Machine Translation (NMT) but can be detrimental for hard-drive prognosis as we will see in the next sections.

\subsubsection{Positional Encoding}
In the vanilla Transformer, for each position $pos$, encoding is as follow:
\begin{equation}
PE_{(pos, 2i)} = \sin\left(\frac{pos}{10000^{2i/d_{\text{model}}}}\right)
\end{equation}

\begin{equation}
PE_{(pos, 2i+1)} = \cos\left(\frac{pos}{10000^{2i/d_{\text{model}}}}\right)
\end{equation}

\subsubsection{Multi-head Attention}
This multi-head attention mechanism allows the model to attend to different parts of the input sequence simultaneously, capturing various patterns and relationships within the data. 

\begin{figure}[htbp]
\centering
\begin{subfigure}[b]{\columnwidth}
        \centering
        \includegraphics[width=\columnwidth]{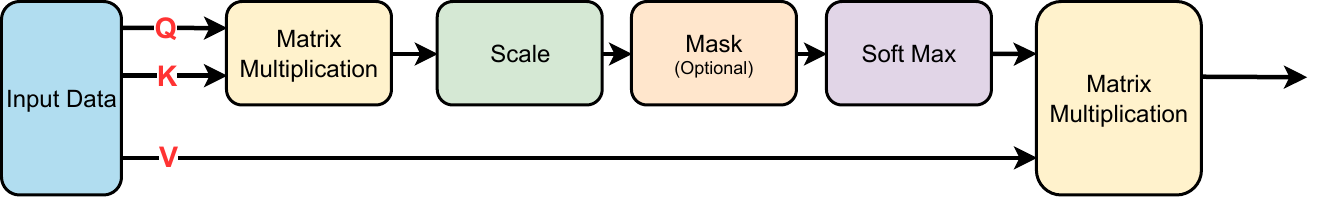}
   \caption{Scaled Dot-Product Self-Attention mechanism}
   \label{fig:scaleddot}
\end{subfigure}
\\~\\
\begin{subfigure}[b]{\columnwidth}
\centering
   \includegraphics[width=\columnwidth]{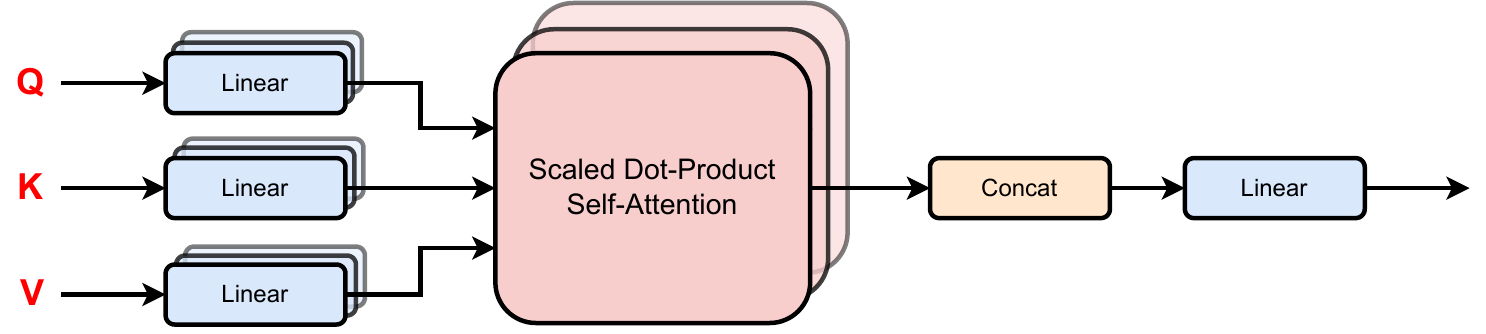}
   \caption{Multi-Head Self-Attention mechanism}
   \label{fig:multihead}
\end{subfigure}
\caption{The process of Multi-Head Self-Attention in Transformers }
\label{fig:Ng2}
\end{figure}

The transformer produces the self attention as visualized in Fig. \ref{fig:scaleddot}. The self-attention can be be formulated as:
\begin{equation}
    \text{Attention}(Q, K, V) = \text{softmax} (\frac{Q K'}{\sqrt{d_{model}}}) V
\end{equation}

We will delve into what $Q$, $K$ and $V$ mean:
\begin{itemize}
    \item \textbf{Query (Q)}: It's the representation of a token (word or position) that you want to calculate the attention scores for concerning other tokens in the sequence. These query vectors are used to determine how much attention each token should pay to other tokens in the sequence.
    \item \textbf{Key (K)}: It's the representation of a token that provides context for computing the attention scores. These key vectors capture information about the other tokens in the sequence.
    \item  \textbf{Value (V)}: These value vectors hold the information that is weighted and combined based on the attention scores.
\end{itemize}

The model attends to data from several representational subspaces at various places with multi-head attention. It also exploits parallelism to increase model training and performance.

\begin{equation}
\text{MultiHead}(Q, K, V) = Concat(head_1, head_2, \ldots, head_h) \cdot W^o
\end{equation}

where $ W^o$ is a learnable weight matrix used to linearly combine the outputs of all attention heads and $head_i = \text{Attention}(Q, K, V)$

\subsubsection{Feed-forward network}
The feed-forward network in a Transformer is an important component that captures complicated patterns in the input data. It is made up of numerous fully-connected layers and non-linear activation functions that allow the model to process and convert information gained via attention mechanisms, boosting its ability to learn and express intricate relationships within sequences.

\subsubsection{Encoder-Decoder Structure}
Using self-attention mechanisms, the encoder analyses the input sequence, acquiring contextual information. The decoder then constructs the output sequence, using contextual information from the encoder and previously created tokens to make coherent translations or predictions. Using self-attention mechanisms, the encoder analyses the input sequence, acquiring contextual information. The decoder then constructs the output sequence, using contextual information from the encoder and previously created tokens to make coherent translations or predictions. 

\subsection{DAST: Dual Aspect Self-Attention Transformer}

\begin{figure*}[htbp]
\centering
\includegraphics[width=0.8\textwidth]{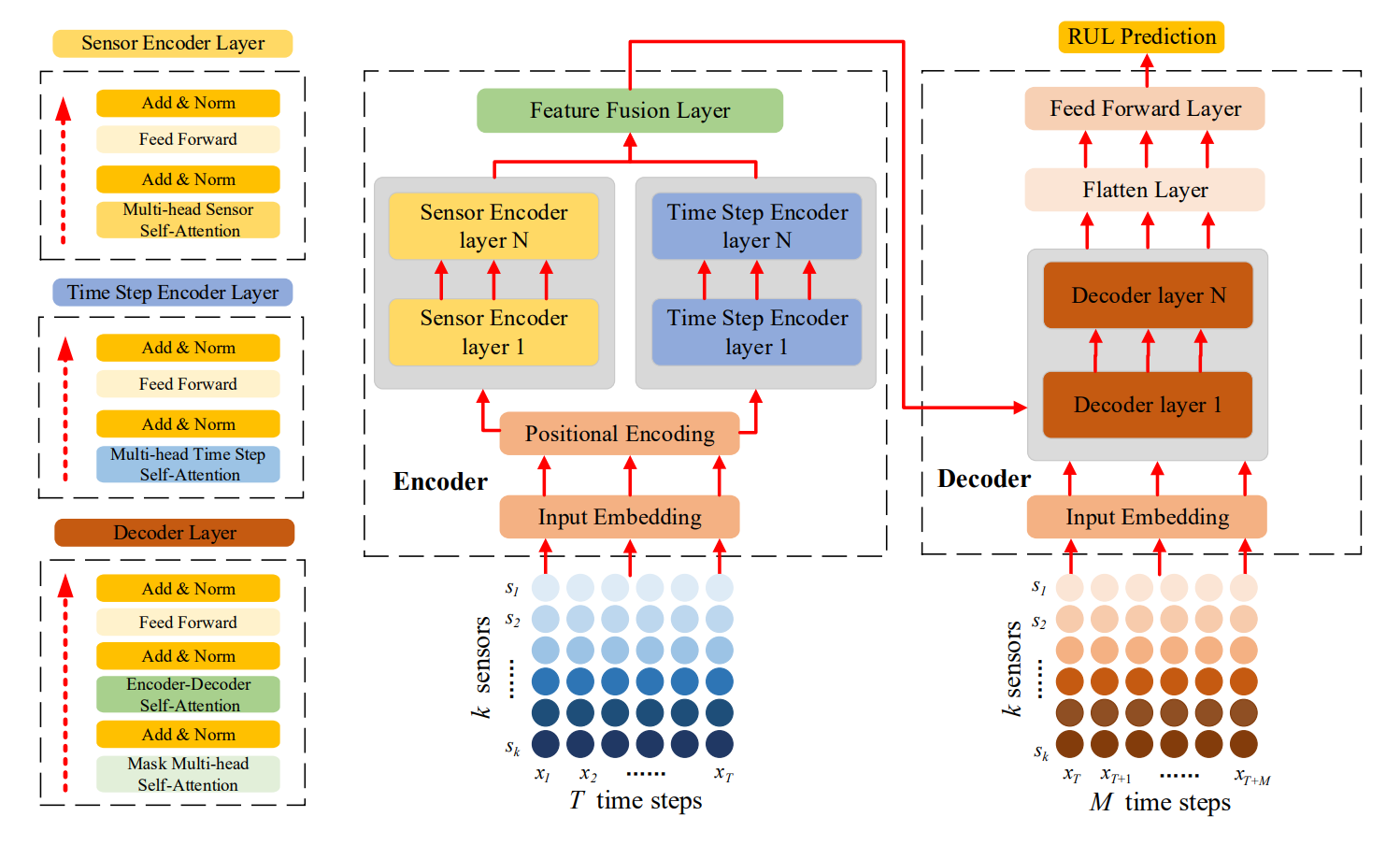}
\caption{Architecture of DAST \cite{dast} }
\label{fig:dast}
\end{figure*}

The DAST \cite{dast} architecture is a recently proposed unique adaptation of the transformer for Turbofan Engine health prognostics on the NASA Commercial Modular Aero-Propulsion System Simulation (CMAPSS) \cite{cmaps} and PHM 2008 data \cite{phm2008}. As shown in Fig. \ref{fig:dast}, it employs two encoders that function in parallel to extract features from various sensors and time steps. DAST encoders are more effective at processing extended data sequences based solely on self-attention and are capable of adaptively learning to focus on more relevant regions of input. Furthermore, the parallel feature extraction approach prevents information from two aspects from influencing each other. 

\section{TFBEST: Temporal-fusion Bi-encoder Self-attention Transformer}
\label{section:tfbest}
Time-stamped data from sensors are the backbone of fault prognosis in industrial electro-mechanical components. It makes possible the identification of temporal patterns, including seasonality or cyclical behaviors, which are crucial for predicting future conditions or occurrences whilst sensors continuously collect data on various aspects of the physical system in real-time. DAST exploits these two features to extract valuable contextual information. TFBEST is a novel transformer that builds on the DAST architecture by replacing the encoding with a learnable positional encoding based on LSTM.

\begin{figure*}[htbp]
\centering
\includegraphics[width=0.8\textwidth]{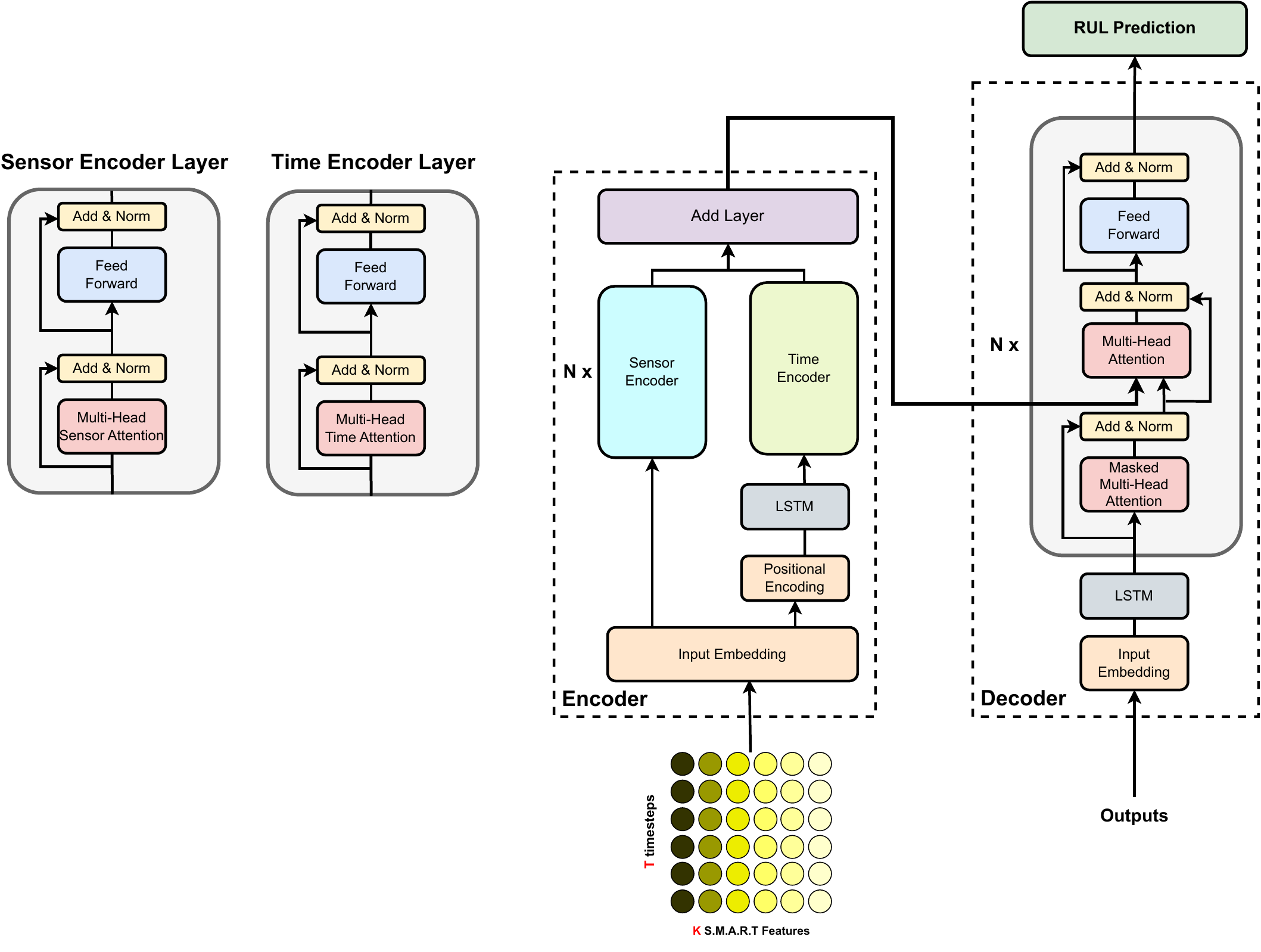}
\caption{Architecture of TFBEST }
\label{fig:tfbest}
\end{figure*}

\subsection{Limitations of absolute positional encoding}
Despite having significant gains in \cite{dast}, absolute position encoding (APE) has limitations for encoding time series data in Transformers. For each time step in the sequence, APE designates a distinct fixed positional embedding. As a result, the model comes to link particular positions with particular embeddings. The significance of various time steps can change in time series data from the actual world, and the fixed embeddings might not adequately capture this dynamic character.  Several studies have revealed that learnable positional embeddings from time series data can be much more effective compared to fixed APE \cite{lim2020temporal, wen2023transformers}. 

\subsection{LSTM based positional encoding}
\label{section:learnablepe}
As highlighted, we use a learnable positional encoding using LSTMs. This allows the model to learn the position of the sensor data across time-steps adaptively during training, making it more flexible and effective at collecting positional information. We also notice a signifact performance gain using this positional encoding compared to APE.

\subsection{Bi-encoders for sensor and time}
Building on the DAST architecture, we employ 2 encoder layers. The sensor encoder applies the multi-head self-attention on the sensor dimension to extract useful information. This layer is an additional layer on the vanilla transformer that focuses on the \SMART features and extracts useful context without feature selection. As shown in Fig. \ref{fig:tfbest}, we also add a time step encoder layer. The time step encoder layer collects features along the time step dimension, allowing the TFBEST model to focus on the time steps that are more essential for RUL prediction. The positional encoding layer processes the time step encoder's input data, which is the transpose of inputs. 

\subsubsection{Sensor Encoder Layer}
The sensor encoder layer is based on vanilla transformer's encoder layer with minor modifications to fit the problem at hand. For every HDD log $i$, we define the senor input across $T$ time-steps as $X_i = \{ X_{1i}, X_{2i}, ..., X_{Ti} \}$. We also define $X'_s$ as the transpose of the inputs passed to the sensor encoder. The  process of self-attention mechanism is visualized in Fig. \ref{fig:Ng2}. We generate the Q, K and V matrices by multiplying $X'_s$ with $W^q_s$, $W^k_s$, and $W^v_s$ (trainable weights) respectively.

\begin{equation}
    Q_s = X'_s W^q_s, \ K_s =  X'_s W^k_s, \ V_s =  X'_s W^v_s
\end{equation}

Then we calculate the dot product of $Q$ and $K$ (scaled by $\sqrt{d_{model}}$), and apply a softmax function along the sensor dimension to obtain the weights of different sensors. Then self-attention can be computed as:
\begin{equation}
    \text{Attention}_s (Q_s, K_s, V_s) = \frac{Q_s K'_s}{\sqrt{d_{model}}} V_s
\end{equation}

We now employ a multi-head attention mechanism. By separating the attention mechanism into many heads, it allows the model to focus on diverse parts of the input sequence at the same time, allowing it to capture complex dependencies and enhance performance on tasks like machine translation and text synthesis. Each head discovers new relationships in the data, improving the model's capacity to handle a wide range of patterns and extract relevant information.

\begin{equation}
    \text{MultiHead} (Q_s, K_s, V_s) = Concat( \{ head_i \}_{i=1}^{h} ) W^o_s
\end{equation}

where $head_i = \text{Attention}_s (Q_s, K_s, V_s)$.

\subsubsection{Time Encoder Layer}
The time encoder layer is very similar to the sensor encoder layer. We directly pass the inputs to a learnable positional encoding as described in section \ref{section:learnablepe}.

\subsubsection{Concatenation of encoder contexts}
Both the sensor encoder and the time step encoder are built by stacking identical sensor or time step encoder layers. For simplicity, we utilize the same number of stacks $N$ for both encoders. The model concatenates information from the two contexts from the encoders. The output from the encoders $O_r$ can be represented:

\begin{equation}
    O_r = Concat(O_s, O_t)
\end{equation}

where $O_s$ and $O_t$ are the context outputs from the sensor and time step encoders. Because the time step encoder and sensor encoder are positioned in parallel, features of the sensor dimension and time step dimension are retrieved concurrently. This approach eliminates the mutual influence of information from the two elements and increases performance by exploiting parallelism.

\subsection{Decoder}
As we have modeled the problem at hand as a regression task, the decoder takes the inputs as is. The decoder block is similar to the one used in \cite{vaswani_trans}. It calculates the source-target attention. An input embedding layer, $n$ decoder layers, an add and normalization layer, and a fully-connected feed forward layer comprise the decoder. The decoder layer consists mostly of two multi-head self-attention sublayers: mask multi-head self-attention and cross-attention sublayer.

\section{Experiments}
\label{section:experiments}
This section introduces a new dataset format, associated experimental settings, and trials on 10-years of Backblaze data to assess TFBEST performance in comparison to cutting-edge RUL prediction techniques and to confirm the benefit of the new architecture.

\subsection{Dataset}
Backblaze is a cloud-based data storage and backup service provider. To support it's day-to-day requirements, it monitors and deploys over 240,000 hard-drives  \cite{q1_2023_stats}. Every quarter, Backblaze publishes daily logs over the quarter. These logs contain information like \SMART features and whether or not a HDD failed on a given day. If it failed, it it marked with a 1 and removed from the subsequent snapshots.

We look at 10-years of worth of data for a particular model of Seagate ST4000DM000 since it is the most failing hard-drive in the cluster \cite{q1_2023_stats}. We build the dataset by looking at different serial numbers of the model. We gather previous 60 days for a particular failing serial number. Then it is concatenated in a sliding window described in the next subsection. For each log, we calculate the RUL. RUL is calculated as follows:

\begin{equation}
    RUL = Failed\ Date - Log\ Creation\ Date
\end{equation}

If a failed device was discovered, for instance, the RUL column would have a value of 1 the day before, 2 the day before that, and so on. We can now treat this as a \textit{regression problem} to predict the RUL given \SMART features.

\subsection{RUL Sequence Generation}
\begin{figure}[htbp]
\centering
\includegraphics[width=0.8\columnwidth]{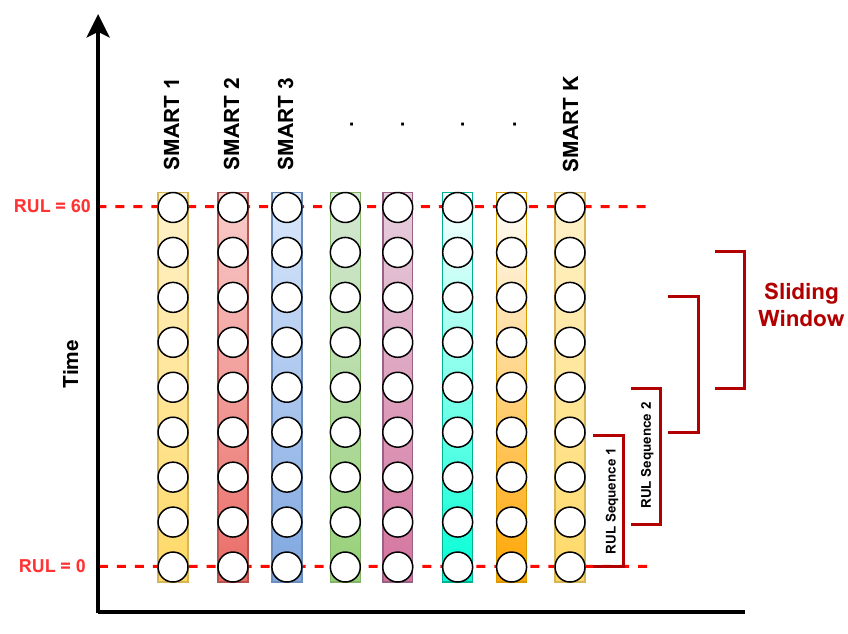}
\caption{RUL Sequence Generation on the Backblaze Dataset }
\label{fig:slidingwindow}
\end{figure}

Using a similar approach in \cite{coursey_lstm}, we employ a different RUL sequence generation compared to \cite{rohan_1} which helps us in providing robust RUL predictions with a confidence interval. The detailed description of the sequence generation is outlined in Fig. \ref{fig:slidingwindow}.

\subsection{Experimental Setting}
The length of the rolling time frame is set at 30. For training, we employ the Adam optimizer with an maximum epochs of 100. The training loss function is RMSE. The learning rate is set to 0.001 with a batch size of 256. Dropout is used for each encoder and decoder layer, with a dropout rate of 0.1. The training loss function can be formulated as:
\begin{equation}
    RMSE = \sqrt{\frac{1}{n}\sum_{i=1}^{n}(Y_i - \hat{Y}_i)^2}
\end{equation}
where $Y_i$ is the actual RUL and $\hat{Y}_i$ is the predicted RUL from the model.

We look at 2013-present data of Seagate ST4000DM000 and build the training, validation and test set as follows:
\begin{itemize}
    \item \textbf{Training}: January 2013 - December 2019
    \item \textbf{Validation}: January 2020 - December 2020
    \item \textbf{Test}: January 2021 - present
\end{itemize}

In this work we employ $h = 4$ parallel attention layers, or heads. Across the models, we use $2$ encoder layers and $1$ decoder layer. For the encoder-decoder LSTM, we use $64$ units LSTM cell and $64$ units feed-forward for the transformers.

\subsection{Comparison to other state-of-the-art models}
Here we compare the performance of TFBEST with state of-the-art deep learning based RUL prediction methods. We consider 3 baselines and compare our model against baselines. The baselines are: 
\begin{itemize}
    \item Encoder-Decoder LSTM
    \item Vanilla Transformer 
    \item Vanilla Transformer with adjusted APE \cite{foumani2023improving}
    \item DAST
\end{itemize}

\begin{table}[htbp]
\centering
\begin{tabular}{|l|c|c|c|}
\hline
\textbf{Model} & \textbf{\begin{tabular}[c]{@{}c@{}}Train\\ RMSE\end{tabular}} & \textbf{\begin{tabular}[c]{@{}c@{}}Validation \\ RMSE\end{tabular}} & \textbf{\begin{tabular}[c]{@{}c@{}}Test \\ RMSE\end{tabular}} \\ \hline
\begin{tabular}[l]{@{}l@{}}Encoder-Decoder \\ LSTM \cite{rohan_1, NIPS2014_a14ac55a}\end{tabular} & 14.46 &  22.19 & 15.25 \\ \hline
\begin{tabular}[l]{@{}l@{}}Vanilla \\ Transformer \cite{vaswani_trans}\end{tabular} & 9.46 & 29.12 & 29.14 \\ \hline
\begin{tabular}[l]{@{}l@{}}Vanilla \\ Transformer with \\ adjusted APE \cite{foumani2023improving} \end{tabular} & 9.44 & 28.23 & 28.22 \\ \hline
DAST \cite{dast} & 9.42 & 13.1 & 13.1 \\ \hline
\textbf{TFBEST} & \textbf{7.75} & \textbf{9.6} & \textbf{9.54} \\ \hline
\end{tabular}
\caption{Performance of TFBEST over other cutting-edge RUL Predictors}
\label{table:progonosis}
\end{table}

\begin{figure}[htbp]
\centering
\begin{subfigure}[b]{1\columnwidth}
        \centering
        \includegraphics[width=1\columnwidth]{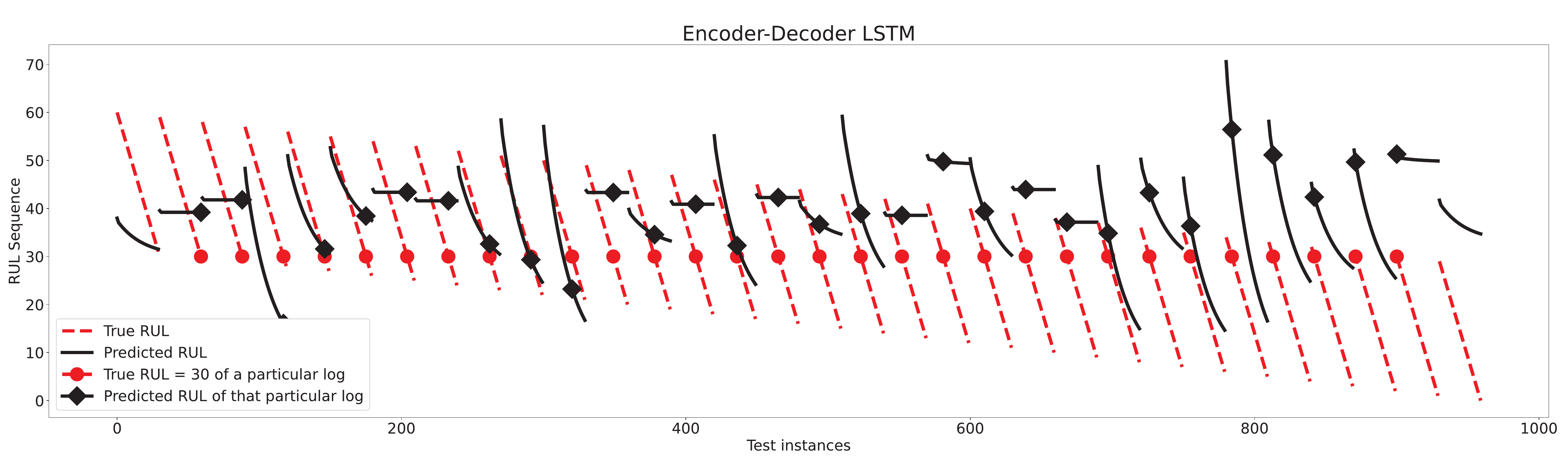}
   \caption{Encoder-Decoder LSTM}
\end{subfigure}

\centering
\begin{subfigure}[b]{1\columnwidth}
        \centering
        \includegraphics[width=\columnwidth]{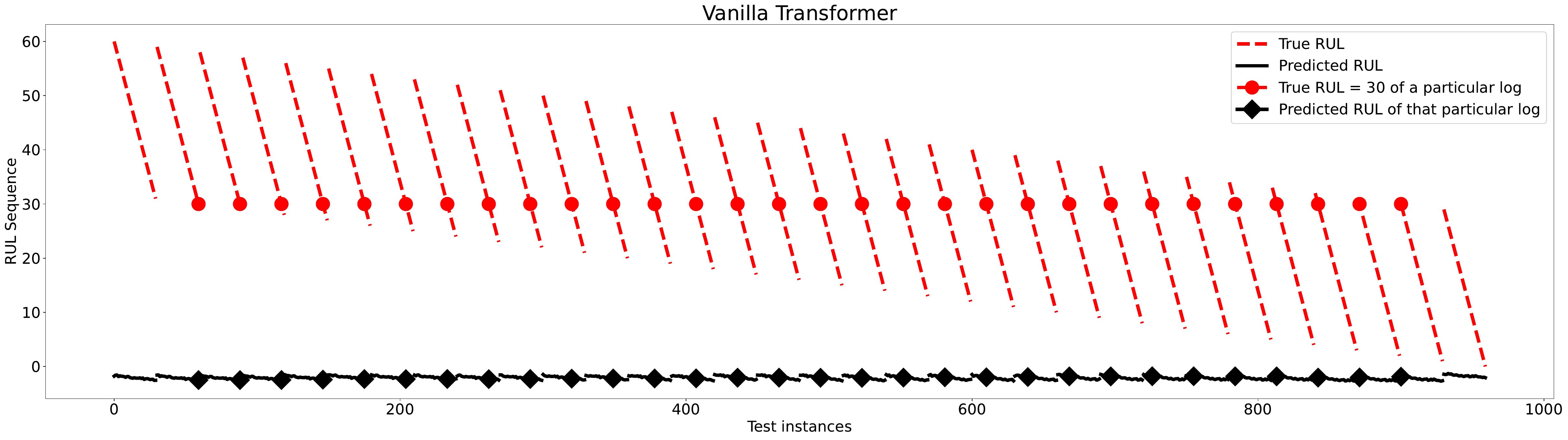}
   \caption{Vanilla Transformer}
\end{subfigure}

\begin{subfigure}[b]{1\columnwidth}
\centering
   \includegraphics[width=\columnwidth]{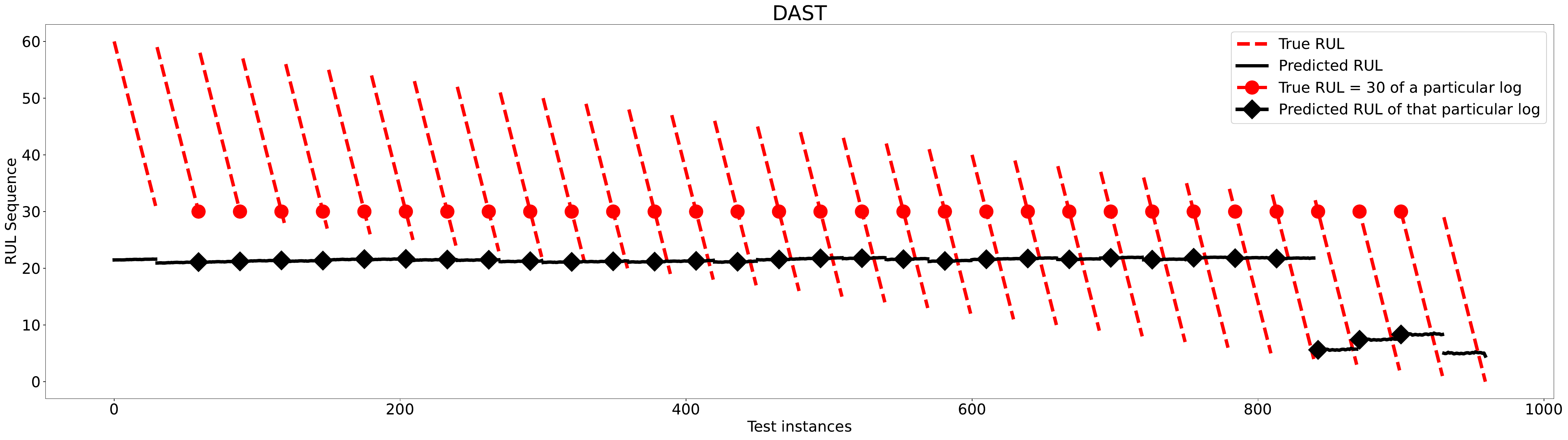}
   \caption{DAST}
\end{subfigure}

\begin{subfigure}[b]{\columnwidth}
\centering
   \includegraphics[width=\columnwidth]{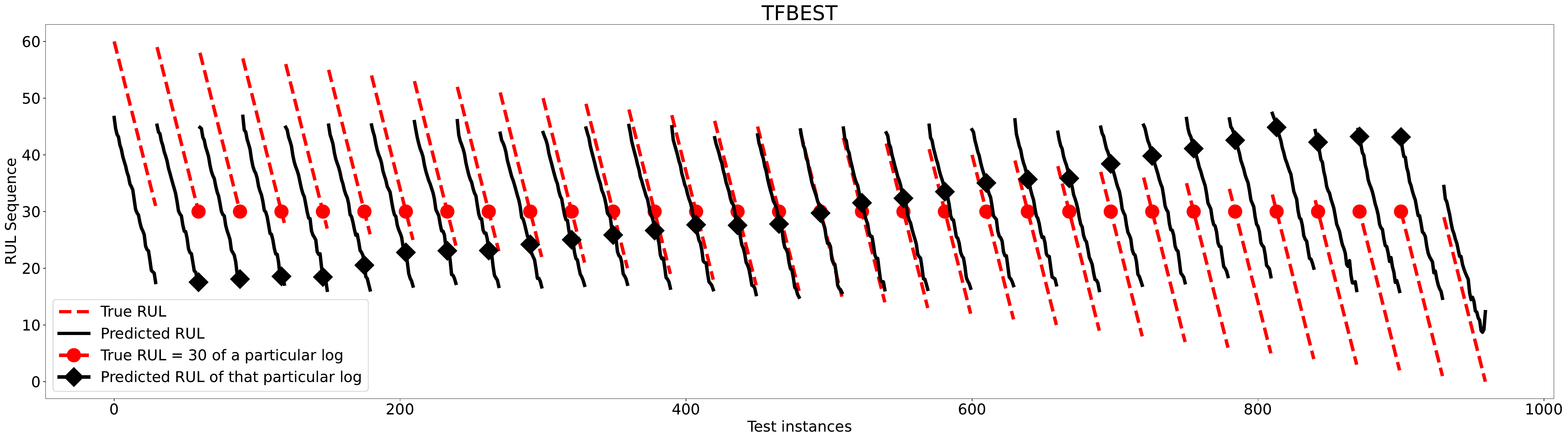}
   \caption{TFBEST}
\end{subfigure}
\caption{Predicted RUL values by different deep-learning models on the Seagate ST4000DM000 model (Serial number: Z305FNVM)}
\label{fig:drive-predictions}
\end{figure}

In Table \ref{table:progonosis}, we can see that LSTMs with feature selection \cite{rohan_1} cannot predict well for long sequences and are very time consuming. It takes approximately 2 minutes per epoch and additional time for pre-processing the data. While transformers are fast for training (\~ 30 seconds per epoch), TFBEST outperforms Vanilla Transformers and DAST by a big margin. This experiment shows the superiority of the model over state-of-the-art architectures. During experimentation, we also can visualize the predictions of one of the failing hard-drive in the Seagate ST4000DM000 models. Fig. \ref{fig:drive-predictions} shows that TFBEST approach understands the underlying sequence in time that is changing based on time steps. The Encoder-Decoder LSTM proves that LSTMs are not suited to handle log sequences whilst predictions by the Vanilla transformer and DAST show that using learnable positional encoding in TFBEST gives near-perfect predictions.

\subsection{Confidence margin statistic }
As we have observed with LSTM-based approaches that are modeled as a regression problem \cite{Anantharaman, coursey_lstm, stackedlstm}, the models output only a single RUL value based on a sequence of data. This may be effective in producing a lower RMSE but confidence from the model may not be adequate. We propose a novel confidence margin metric for RUL prediction. This gives us a error margin and a closed interval of confidence that the RUL will lie in that range.

\begin{equation}
    \text{Confidence Margin} = \hat{\theta} \pm E
\end{equation}

where, $\hat{\theta}$ represents the \textit{point estimate} of the RUL and $E$ represents the \textit{margin-of-error}.

\begin{table*}[]
\centering
\scalebox{1}{\begin{tabular}{|l|ll|ll|ll|ll|}
\hline
\multicolumn{1}{|c|}{\textbf{True RUL}} & \multicolumn{2}{c|}{\textbf{\begin{tabular}[c]{@{}c@{}}Encoder \\ Decoder LSTM\end{tabular}}} & \multicolumn{2}{c|}{\textbf{\begin{tabular}[c]{@{}c@{}}Vanilla \\ Transformer\end{tabular}}} & \multicolumn{2}{c|}{\textbf{DAST}} & \multicolumn{2}{c|}{\textbf{TFBEST}} \\ \hline
\multicolumn{1}{|c|}{} & \multicolumn{1}{c|}{\textbf{\begin{tabular}[c]{@{}c@{}}Point \\ Estimate\end{tabular}}} & \multicolumn{1}{c|}{\textbf{\begin{tabular}[c]{@{}c@{}}90\% \\ Confidence \\ Interval\end{tabular}}} & \multicolumn{1}{c|}{\textbf{\begin{tabular}[c]{@{}c@{}}Point \\ Estimate\end{tabular}}} & \multicolumn{1}{c|}{\textbf{\begin{tabular}[c]{@{}c@{}}90\% \\ Confidence \\ Interval\end{tabular}}} & \multicolumn{1}{c|}{\textbf{\begin{tabular}[c]{@{}c@{}}Point \\ Estimate\end{tabular}}} & \multicolumn{1}{c|}    {\textbf{\begin{tabular}[c]{@{}c@{}}90\% \\ Confidence \\ Interval\end{tabular}}} & \multicolumn{1}{c|}{\textbf{\begin{tabular}[c]{@{}c@{}}Point \\ Estimate\end{tabular}}} & \multicolumn{1}{c|}{\textbf{\begin{tabular}[c]{@{}c@{}}90\% \\ Confidence \\ Interval\end{tabular}}} \\ \hline
\textbf{} & \multicolumn{1}{l|}{} &  & \multicolumn{1}{l|}{} &  & \multicolumn{1}{l|}{} &  & \multicolumn{1}{l|}{} &  \\ \hline
\textbf{60} & \multicolumn{1}{l|}{37.98  $\pm$  0.00} & (37.98, 37.98) & \multicolumn{1}{l|}{3.72 $\pm$ 0.01} & (3.73, 3.87) & \multicolumn{1}{l|}{29.98  $\pm$  0.00} & (29.98, 29.98) & \multicolumn{1}{l|}{45.13  $\pm$  0.00} & (45.13, 45.13) \\ \hline
\textbf{59} & \multicolumn{1}{l|}{38.31  $\pm$  1.24} & (30.46, 46.15) & \multicolumn{1}{l|}{3.80  $\pm$  0.01} & (3.73, 3.87) & \multicolumn{1}{l|}{29.96  $\pm$  0.01} & (29.92, 30.00) & \multicolumn{1}{l|}{44.42  $\pm$  0.29} & (42.59, 46.26) \\ \hline
\textbf{58} & \multicolumn{1}{l|}{39.36  $\pm$  1.59} & (34.73, 44.00) & \multicolumn{1}{l|}{3.82  $\pm$  0.01} & (3.80, 3.84) & \multicolumn{1}{l|}{29.99  $\pm$  0.07} & (29.77, 30.20) & \multicolumn{1}{l|}{43.40  $\pm$  0.28} & (42.59, 44.22) \\ \hline
\textbf{57} & \multicolumn{1}{l|}{41.42  $\pm$  2.56} & (35.39, 47.45) & \multicolumn{1}{l|}{3.71  $\pm$  0.01} & (3.70, 3.73) & \multicolumn{1}{l|}{29.91  $\pm$  0.09} & (29.70, 30.12) & \multicolumn{1}{l|}{43.57  $\pm$  0.25} & (42.99, 44.15) \\ \hline
\textbf{56} & \multicolumn{1}{l|}{42.69  $\pm$  2.59} & (37.17, 48.21) & \multicolumn{1}{l|}{3.79  $\pm$  0.00} & (3.78, 3.79) & \multicolumn{1}{l|}{29.92  $\pm$  0.06} & (29.80, 30.04) & \multicolumn{1}{l|}{43.34  $\pm$  0.68} & (41.89, 44.79) \\ \hline
\textbf{55} & \multicolumn{1}{l|}{43.61  $\pm$  2.55} & (38.47, 48.76) & \multicolumn{1}{l|}{3.77  $\pm$  0.01} & (3.76, 3.79) & \multicolumn{1}{l|}{29.89  $\pm$  0.07} & (29.76, 30.03) & \multicolumn{1}{l|}{42.74  $\pm$  0.77} & (41.19, 44.29) \\ \hline
\textbf{54} & \multicolumn{1}{l|}{42.92  $\pm$  1.96} & (39.10, 46.73) & \multicolumn{1}{l|}{3.79  $\pm$  0.01} & (3.77, 3.80) & \multicolumn{1}{l|}{29.88  $\pm$  0.07} & (29.75, 30.02) & \multicolumn{1}{l|}{42.29  $\pm$  0.78} & (40.77, 43.80) \\ \hline
\textbf{53} & \multicolumn{1}{l|}{42.20  $\pm$  1.62} & (39.12, 45.28) & \multicolumn{1}{l|}{3.77  $\pm$  0.01} & (3.76, 3.78) & \multicolumn{1}{l|}{29.84  $\pm$  0.08} & (29.68, 29.99) & \multicolumn{1}{l|}{41.57  $\pm$  0.74} & (40.16, 42.98) \\ \hline
\textbf{52} & \multicolumn{1}{l|}{42.42  $\pm$  1.59} & (39.46, 45.39) & \multicolumn{1}{l|}{3.72  $\pm$  0.01} & (3.71, 3.73) & \multicolumn{1}{l|}{29.75  $\pm$  0.09} & (29.58, 29.91) & \multicolumn{1}{l|}{40.71  $\pm$  0.76} & (39.30, 42.12) \\ \hline
\textbf{51} & \multicolumn{1}{l|}{43.46  $\pm$  2.15} & (39.52, 47.40) & \multicolumn{1}{l|}{3.77  $\pm$  0.01} & (3.76, 3.78) & \multicolumn{1}{l|}{29.77  $\pm$  0.07} & (29.64, 29.90) & \multicolumn{1}{l|}{39.75  $\pm$  0.98} & (37.96, 41.54) \\ \hline
\textbf{50} & \multicolumn{1}{l|}{43.99  $\pm$  2.20} & (40.01, 47.97) & \multicolumn{1}{l|}{3.80  $\pm$  0.01} & (3.79, 3.80) & \multicolumn{1}{l|}{29.78  $\pm$  0.08} & (29.64, 29.93) & \multicolumn{1}{l|}{39.53  $\pm$  1.09} & (37.55, 41.51) \\ \hline
\textbf{49} & \multicolumn{1}{l|}{43.14  $\pm$  1.80} & (39.90, 46.38) & \multicolumn{1}{l|}{3.72  $\pm$  0.01} & (3.71, 3.73) & \multicolumn{1}{l|}{29.67  $\pm$  0.10} & (29.49, 29.85) & \multicolumn{1}{l|}{39.61  $\pm$  1.11} & (37.62, 41.59) \\ \hline
\textbf{48} & \multicolumn{1}{l|}{42.19  $\pm$  1.55} & (39.42, 44.95) & \multicolumn{1}{l|}{3.72  $\pm$  0.01} & (3.70, 3.73) & \multicolumn{1}{l|}{29.54  $\pm$  0.13} & (29.31, 29.77) & \multicolumn{1}{l|}{38.94  $\pm$  1.17} & (36.86, 41.02) \\ \hline
\textbf{47} & \multicolumn{1}{l|}{41.48  $\pm$  1.36} & (39.07, 43.89) & \multicolumn{1}{l|}{3.69  $\pm$  0.01} & (3.68, 3.70) & \multicolumn{1}{l|}{29.49  $\pm$  0.13} & (29.25, 29.73) & \multicolumn{1}{l|}{38.26  $\pm$  1.24} & (36.06, 40.46) \\ \hline
\textbf{46} & \multicolumn{1}{l|}{41.81  $\pm$  1.54} & (39.09, 44.53) & \multicolumn{1}{l|}{3.73  $\pm$  0.01} & (3.72, 3.75) & \multicolumn{1}{l|}{29.55  $\pm$  0.10} & (29.37, 29.74) & \multicolumn{1}{l|}{37.77  $\pm$  1.32} & (35.44, 40.10) \\ \hline
\textbf{45} & \multicolumn{1}{l|}{41.21  $\pm$  1.35} & (38.86, 43.57) & \multicolumn{1}{l|}{3.71  $\pm$  0.01} & (3.70, 3.73) & \multicolumn{1}{l|}{29.48  $\pm$  0.12} & (29.28, 29.69) & \multicolumn{1}{l|}{37.51  $\pm$  1.28} & (35.27, 39.75) \\ \hline
\textbf{44} & \multicolumn{1}{l|}{40.66  $\pm$  1.22} & (38.52, 42.79) & \multicolumn{1}{l|}{3.73  $\pm$  0.01} & (3.72, 3.74) & \multicolumn{1}{l|}{29.45  $\pm$  0.13} & (29.21, 29.68) & \multicolumn{1}{l|}{37.04  $\pm$  1.31} & (34.75, 39.32) \\ \hline
\textbf{43} & \multicolumn{1}{l|}{41.15  $\pm$  1.55} & (38.45, 43.85) & \multicolumn{1}{l|}{3.75  $\pm$  0.01} & (3.74, 3.77) & \multicolumn{1}{l|}{29.40  $\pm$  0.16} & (29.13, 29.67) & \multicolumn{1}{l|}{36.56  $\pm$  1.33} & (34.24, 38.88) \\ \hline
\textbf{42} & \multicolumn{1}{l|}{40.43  $\pm$  1.39} & (38.02, 42.84) & \multicolumn{1}{l|}{3.78  $\pm$  0.01} & (3.76, 3.79) & \multicolumn{1}{l|}{29.44  $\pm$  0.13} & (29.22, 29.67) & \multicolumn{1}{l|}{35.80  $\pm$  1.46} & (33.26, 38.33) \\ \hline
\textbf{41} & \multicolumn{1}{l|}{40.43  $\pm$  1.40} & (38.00, 42.85) & \multicolumn{1}{l|}{3.76  $\pm$  0.01} & (3.74, 3.77) & \multicolumn{1}{l|}{29.21  $\pm$  0.19} & (28.88, 29.55) & \multicolumn{1}{l|}{35.50  $\pm$  1.39} & (33.11, 37.90) \\ \hline
\textbf{40} & \multicolumn{1}{l|}{40.40  $\pm$  1.40} & (37.98, 42.83) & \multicolumn{1}{l|}{3.79  $\pm$  0.01} & (3.78, 3.80) & \multicolumn{1}{l|}{29.39  $\pm$  0.15} & (29.13, 29.65) & \multicolumn{1}{l|}{34.69  $\pm$  1.59} & (31.95, 37.43) \\ \hline
\textbf{39} & \multicolumn{1}{l|}{40.08  $\pm$  1.34} & (37.78, 42.38) & \multicolumn{1}{l|}{3.68  $\pm$  0.01} & (3.67, 3.69) & \multicolumn{1}{l|}{28.99  $\pm$  0.22} & (28.61, 29.37) & \multicolumn{1}{l|}{34.59  $\pm$  1.56} & (31.91, 37.27) \\ \hline
\textbf{38} & \multicolumn{1}{l|}{39.52  $\pm$  1.28} & (37.32, 41.73) & \multicolumn{1}{l|}{3.73  $\pm$  0.01} & (3.72, 3.74) & \multicolumn{1}{l|}{29.15  $\pm$  0.18} & (28.85, 29.45) & \multicolumn{1}{l|}{34.41  $\pm$  1.63} & (31.61, 37.22) \\ \hline
\textbf{37} & \multicolumn{1}{l|}{39.50  $\pm$  1.31} & (37.26, 41.74) & \multicolumn{1}{l|}{3.77  $\pm$  0.01} & (3.76, 3.77) & \multicolumn{1}{l|}{29.13  $\pm$  0.18} & (28.81, 29.44) & \multicolumn{1}{l|}{34.03  $\pm$  1.70} & (31.12, 36.94) \\ \hline
\textbf{36} & \multicolumn{1}{l|}{39.46  $\pm$  1.33} & (37.19, 41.73) & \multicolumn{1}{l|}{3.76  $\pm$  0.01} & (3.75, 3.77) & \multicolumn{1}{l|}{29.00  $\pm$  0.20} & (28.66, 29.35) & \multicolumn{1}{l|}{33.91  $\pm$  1.70} & (30.99, 36.82) \\ \hline
\textbf{35} & \multicolumn{1}{l|}{39.25  $\pm$  1.30} & (37.03, 41.46) & \multicolumn{1}{l|}{3.71  $\pm$  0.01} & (3.70, 3.73) & \multicolumn{1}{l|}{28.73  $\pm$  0.25} & (28.31, 29.16) & \multicolumn{1}{l|}{33.22  $\pm$  1.66} & (30.39, 36.06) \\ \hline
\textbf{34} & \multicolumn{1}{l|}{39.90  $\pm$  1.71} & (36.97, 42.82) & \multicolumn{1}{l|}{3.74  $\pm$  0.01} & (3.73, 3.76) & \multicolumn{1}{l|}{28.78  $\pm$  0.24} & (28.37, 29.19) & \multicolumn{1}{l|}{32.88  $\pm$  1.73} & (29.93, 35.83) \\ \hline
\textbf{33} & \multicolumn{1}{l|}{39.94  $\pm$  1.70} & (37.04, 42.83) & \multicolumn{1}{l|}{3.75  $\pm$  0.01} & (3.74, 3.76) & \multicolumn{1}{l|}{28.81  $\pm$  0.22} & (28.44, 29.19) & \multicolumn{1}{l|}{32.31  $\pm$  1.80} & (29.26, 35.37) \\ \hline
\textbf{32} & \multicolumn{1}{l|}{39.50  $\pm$  1.58} & (36.81, 42.18) & \multicolumn{1}{l|}{3.68  $\pm$  0.01} & (3.67, 3.69) & \multicolumn{1}{l|}{28.11  $\pm$  0.33} & (27.55, 28.66) & \multicolumn{1}{l|}{31.87  $\pm$  1.75} & (28.89, 34.84) \\ \hline
\textbf{31} & \multicolumn{1}{l|}{39.32  $\pm$  1.54} & (36.70, 41.93) & \multicolumn{1}{l|}{3.70  $\pm$  0.01} & (3.69, 3.71) & \multicolumn{1}{l|}{28.00  $\pm$  0.32} & (27.44, 28.55) & \multicolumn{1}{l|}{30.21  $\pm$  1.62} & (27.46, 32.96) \\ \hline
\textbf{30} & \multicolumn{1}{l|}{39.34  $\pm$  1.52} & (36.76, 41.93) & \multicolumn{1}{l|}{3.72  $\pm$  0.01} & (3.71, 3.73) & \multicolumn{1}{l|}{28.07  $\pm$  0.29} & (27.58, 28.57) & \multicolumn{1}{l|}{29.92  $\pm$  1.55} & (27.29, 32.55) \\ \hline
\textbf{29} & \multicolumn{1}{l|}{38.83  $\pm$  1.50} & (36.29, 41.37) & \multicolumn{1}{l|}{3.79  $\pm$  0.07} & (3.67, 3.91) & \multicolumn{1}{l|}{27.35  $\pm$  0.45} & (26.58, 28.12) & \multicolumn{1}{l|}{29.77  $\pm$  1.46} & (27.30, 32.25) \\ \hline
\textbf{28} & \multicolumn{1}{l|}{38.13  $\pm$  1.53} & (35.52, 40.73) & \multicolumn{1}{l|}{3.76  $\pm$  0.08} & (3.63, 3.89) & \multicolumn{1}{l|}{27.41  $\pm$  0.45} & (26.64, 28.18) & \multicolumn{1}{l|}{28.62  $\pm$  1.42} & (26.20, 31.03) \\ \hline
\textbf{27} & \multicolumn{1}{l|}{38.40  $\pm$  1.33} & (36.13, 40.67) & \multicolumn{1}{l|}{3.80  $\pm$  0.08} & (3.66, 3.93) & \multicolumn{1}{l|}{27.38  $\pm$  0.46} & (26.59, 28.17) & \multicolumn{1}{l|}{28.05  $\pm$  1.38} & (25.71, 30.39) \\ \hline
\textbf{26} & \multicolumn{1}{l|}{38.15  $\pm$  1.36} & (35.83, 40.48) & \multicolumn{1}{l|}{3.90  $\pm$  0.08} & (3.76, 4.04) & \multicolumn{1}{l|}{27.64  $\pm$  0.46} & (26.85, 28.43) & \multicolumn{1}{l|}{27.75  $\pm$  1.46} & (25.27, 30.23) \\ \hline
\textbf{25} & \multicolumn{1}{l|}{37.66  $\pm$  1.44} & (35.21, 40.12) & \multicolumn{1}{l|}{3.82  $\pm$  0.08} & (3.68, 3.97) & \multicolumn{1}{l|}{27.60  $\pm$  0.47} & (26.79, 28.41) & \multicolumn{1}{l|}{27.49  $\pm$  1.39} & (25.11, 29.86) \\ \hline
\textbf{24} & \multicolumn{1}{l|}{36.93  $\pm$  1.50} & (34.36, 39.50) & \multicolumn{1}{l|}{3.87  $\pm$  0.09} & (3.72, 4.02) & \multicolumn{1}{l|}{27.63  $\pm$  0.49} & (26.79, 28.47) & \multicolumn{1}{l|}{27.09  $\pm$  1.35} & (24.79, 29.39) \\ \hline
\textbf{23} & \multicolumn{1}{l|}{36.23  $\pm$  1.59} & (33.52, 38.95) & \multicolumn{1}{l|}{3.81  $\pm$  0.09} & (3.65, 3.97) & \multicolumn{1}{l|}{27.11  $\pm$  0.51} & (26.24, 27.99) & \multicolumn{1}{l|}{26.08  $\pm$  1.25} & (23.95, 28.22) \\ \hline
\textbf{22} & \multicolumn{1}{l|}{36.00  $\pm$  1.67} & (33.12, 38.87) & \multicolumn{1}{l|}{3.76  $\pm$  0.10} & (3.59, 3.92) & \multicolumn{1}{l|}{27.28  $\pm$  0.52} & (26.38, 28.18) & \multicolumn{1}{l|}{25.63  $\pm$  1.20} & (23.57, 27.69) \\ \hline
\textbf{21} & \multicolumn{1}{l|}{36.05  $\pm$  1.72} & (33.10, 39.00) & \multicolumn{1}{l|}{3.79  $\pm$  0.10} & (3.62, 3.97) & \multicolumn{1}{l|}{27.22  $\pm$  0.55} & (26.28, 28.15) & \multicolumn{1}{l|}{25.36  $\pm$  1.24} & (23.22, 27.50) \\ \hline
\textbf{20} & \multicolumn{1}{l|}{36.54  $\pm$  1.58} & (33.82, 39.26) & \multicolumn{1}{l|}{3.82  $\pm$  0.11} & (3.64, 4.01) & \multicolumn{1}{l|}{27.09  $\pm$  0.57} & (26.11, 28.07) & \multicolumn{1}{l|}{24.85  $\pm$  1.25} & (22.69, 27.00) \\ \hline
\textbf{19} & \multicolumn{1}{l|}{35.78  $\pm$  1.68} & (32.88, 38.68) & \multicolumn{1}{l|}{3.78  $\pm$  0.11} & (3.59, 3.98) & \multicolumn{1}{l|}{26.60  $\pm$  0.59} & (25.57, 27.62) & \multicolumn{1}{l|}{23.54  $\pm$  1.14} & (21.57, 25.51) \\ \hline
\textbf{18} & \multicolumn{1}{l|}{35.49  $\pm$  1.83} & (32.32, 38.66) & \multicolumn{1}{l|}{3.76  $\pm$  0.12} & (3.56, 3.97) & \multicolumn{1}{l|}{26.25  $\pm$  0.63} & (25.17, 27.34) & \multicolumn{1}{l|}{23.02  $\pm$  1.10} & (21.11, 24.93) \\ \hline
\textbf{17} & \multicolumn{1}{l|}{34.77  $\pm$  1.97} & (31.34, 38.20) & \multicolumn{1}{l|}{3.83  $\pm$  0.13} & (3.61, 4.04) & \multicolumn{1}{l|}{26.57  $\pm$  0.66} & (25.42, 27.72) & \multicolumn{1}{l|}{23.10  $\pm$  1.14} & (21.12, 25.09) \\ \hline
\textbf{16} & \multicolumn{1}{l|}{34.99  $\pm$  2.06} & (31.40, 38.58) & \multicolumn{1}{l|}{3.79  $\pm$  0.13} & (3.56, 4.02) & \multicolumn{1}{l|}{26.00  $\pm$  0.69} & (24.79, 27.21) & \multicolumn{1}{l|}{21.86  $\pm$  1.15} & (19.85, 23.88) \\ \hline
\textbf{15} & \multicolumn{1}{l|}{34.13  $\pm$  2.21} & (30.26, 38.00) & \multicolumn{1}{l|}{3.77  $\pm$  0.14} & (3.52, 4.02) & \multicolumn{1}{l|}{25.54  $\pm$  0.72} & (24.28, 26.81) & \multicolumn{1}{l|}{21.77  $\pm$  1.15} & (19.76, 23.78) \\ \hline
\textbf{14} & \multicolumn{1}{l|}{33.70  $\pm$  2.43} & (29.41, 37.98) & \multicolumn{1}{l|}{3.85  $\pm$  0.15} & (3.58, 4.12) & \multicolumn{1}{l|}{25.80  $\pm$  0.77} & (24.44, 27.16) & \multicolumn{1}{l|}{20.87  $\pm$  1.24} & (18.69, 23.05) \\ \hline
\textbf{13} & \multicolumn{1}{l|}{33.73  $\pm$  2.65} & (29.03, 38.43) & \multicolumn{1}{l|}{3.89  $\pm$  0.16} & (3.60, 4.17) & \multicolumn{1}{l|}{25.51  $\pm$  0.82} & (24.06, 26.97) & \multicolumn{1}{l|}{20.20  $\pm$  1.29} & (17.91, 22.49) \\ \hline
\textbf{12} & \multicolumn{1}{l|}{32.97  $\pm$  2.91} & (27.78, 38.16) & \multicolumn{1}{l|}{3.91  $\pm$  0.18} & (3.60, 4.22) & \multicolumn{1}{l|}{25.63  $\pm$  0.90} & (24.03, 27.23) & \multicolumn{1}{l|}{20.00  $\pm$  1.32} & (17.64, 22.36) \\ \hline
\textbf{11} & \multicolumn{1}{l|}{31.22  $\pm$  2.87} & (26.06, 36.37) & \multicolumn{1}{l|}{3.84  $\pm$  0.19} & (3.50, 4.18) & \multicolumn{1}{l|}{24.28  $\pm$  0.94} & (22.60, 25.96) & \multicolumn{1}{l|}{18.86  $\pm$  1.51} & (16.15, 21.58) \\ \hline
\textbf{10} & \multicolumn{1}{l|}{30.92  $\pm$  3.22} & (25.08, 36.76) & \multicolumn{1}{l|}{3.90  $\pm$  0.21} & (3.53, 4.28) & \multicolumn{1}{l|}{23.99  $\pm$  1.01} & (22.17, 25.82) & \multicolumn{1}{l|}{17.44  $\pm$  1.59} & (14.55, 20.32) \\ \hline
\textbf{9} & \multicolumn{1}{l|}{29.22  $\pm$  3.33} & (23.12, 35.32) & \multicolumn{1}{l|}{3.99  $\pm$  0.23} & (3.57, 4.41) & \multicolumn{1}{l|}{24.36  $\pm$  1.12} & (22.31, 26.41) & \multicolumn{1}{l|}{17.06  $\pm$  1.92} & (13.54, 20.57) \\ \hline
\textbf{8} & \multicolumn{1}{l|}{27.93  $\pm$  3.66} & (21.13, 34.73) & \multicolumn{1}{l|}{3.96  $\pm$  0.25} & (3.48, 4.43) & \multicolumn{1}{l|}{23.27  $\pm$  1.20} & (21.03, 25.51) & \multicolumn{1}{l|}{15.62  $\pm$  1.89} & (12.10, 19.13) \\ \hline
\textbf{7} & \multicolumn{1}{l|}{29.16  $\pm$  3.80} & (21.96, 36.35) & \multicolumn{1}{l|}{4.10  $\pm$  0.29} & (3.56, 4.64) & \multicolumn{1}{l|}{23.19  $\pm$  1.32} & (20.68, 25.70) & \multicolumn{1}{l|}{14.67  $\pm$  2.12} & (10.64, 18.69) \\ \hline
\textbf{6} & \multicolumn{1}{l|}{28.40  $\pm$  4.45} & (19.76, 37.05) & \multicolumn{1}{l|}{4.09  $\pm$  0.33} & (3.45, 4.73) & \multicolumn{1}{l|}{22.57  $\pm$  1.52} & (19.62, 25.51) & \multicolumn{1}{l|}{13.24  $\pm$  2.41} & (8.55, 17.94) \\ \hline
\textbf{5} & \multicolumn{1}{l|}{30.34  $\pm$  4.62} & (21.04, 39.64) & \multicolumn{1}{l|}{4.10  $\pm$  0.38} & (3.33, 4.87) & \multicolumn{1}{l|}{21.08  $\pm$  1.63} & (17.79, 24.37) & \multicolumn{1}{l|}{11.06  $\pm$  2.40} & (6.22, 15.90) \\ \hline
\textbf{4} & \multicolumn{1}{l|}{32.83  $\pm$  4.62} & (22.98, 42.68) & \multicolumn{1}{l|}{4.19  $\pm$  0.46} & (3.20, 5.17) & \multicolumn{1}{l|}{19.93  $\pm$  1.65} & (16.41, 23.44) & \multicolumn{1}{l|}{8.79  $\pm$  2.35} & (3.78, 13.80) \\ \hline
\textbf{3} & \multicolumn{1}{l|}{34.63  $\pm$  5.46} & (21.78, 47.48) & \multicolumn{1}{l|}{4.31  $\pm$  0.58} & (2.95, 5.67) & \multicolumn{1}{l|}{13.98  $\pm$  0.54} & (12.72, 15.24) & \multicolumn{1}{l|}{5.53  $\pm$  0.40} & (4.59, 6.48) \\ \hline
\textbf{2} & \multicolumn{1}{l|}{36.79  $\pm$  7.09} & (16.09, 57.48) & \multicolumn{1}{l|}{4.44  $\pm$  0.77} & (2.19, 6.70) & \multicolumn{1}{l|}{14.86  $\pm$  0.59} & (13.15, 16.57) & \multicolumn{1}{l|}{4.99  $\pm$  0.23} & (4.32, 5.65) \\ \hline
\textbf{1} & \multicolumn{1}{l|}{42.34  $\pm$  7.54} & (-5.29, 89.97) & \multicolumn{1}{l|}{4.88  $\pm$  1.16} & (-2.47, 12.22) & \multicolumn{1}{l|}{15.43  $\pm$  0.88} & (9.88, 20.97) & \multicolumn{1}{l|}{4.23  $\pm$  0.07} & (3.78, 4.69) \\ \hline
\textbf{0} & \multicolumn{1}{l|}{34.71  $\pm$  0.00} & (nan, nan) & \multicolumn{1}{l|}{6.03  $\pm$  0.00} & (nan, nan) & \multicolumn{1}{l|}{6.93  $\pm$  0.00} & (nan, nan) & \multicolumn{1}{l|}{3.72  $\pm$  0.00} & (nan, nan) \\ \hline
\end{tabular}}
\caption{Confidence interval evaluation on different deep-learning models on the Seagate ST4000DM000 model (Serial number: Z305FNVM)}
\label{table:confidencei}
\end{table*}

The confidence margin is calculated as we use a RUL sequence with overlapping values. These overlapping sequences can give a confidence-based estimate of the RUL. A point estimate, on the other hand can be misleading if the model is not perfect. As we can see from the Table \ref{table:confidencei}, TFBEST produces near perfect point estimates with lower margin-of-error compared to the rest of the models.

\section{Discussion}
We propose TFBEST, a novel transformer architecture for prediction of of hard-drive failures and highlight its advantages over the vanilla Transformer and DAST. To begin with, existing deep learning-based RUL prediction approaches rely heavily on the RNN/CNN architecture. While the use of LSTM has provided very good results in previous works, most architectures have to depend on careful feature engineering to extract the useful \SMART features. In contrast to preceding architectures, our solution is based on the Transformer architecture, which relies solely on the self-attention mechanism to analyze \SMART features. Our architecture is also superior to the DAST architecture by incorporating time-step positions via a learnable encoding technique. This has improved performance of predictions significantly. 

We also use a new confidence-margin interval to provide a more reliable RUL estimate. RUL predictions can be very far from real truth if we only rely on a single output from the model. It is also more useful to predict a range of RUL for a particular log so that the datacenter can evaluate when to change the drive in a timeframe. Compared to previous literature, where either deep-learning models are used as a binary classifier or predicting a single value, our metric is inspired by confidence around failure. We rely on RUL sequences which create an overlap across 60 days by a rolling window. Experimental studies with the above RUL prediction method validate the advantage of our method and model.

\section{Conclusion and Future Work}
\label{section:future}

We propose TFBEST, a novel transformer architecture for 
HDD RUL estimation. TFBEST uses a learnable positional encoding based on LSTMs and the self-attention mechanism to process the whole sequence of CBM data. It uses a sensor encoder and a time step encoder to simultaneously record the weighted characteristics of thee data. 
The TFBEST model adaptively learns the significance of various sensors and time steps without the requirement for feature selection through attention mechanism. The performance of our method for RUL prediction is better than state-of-the-art deep RUL prediction methods, according to experimental findings on actual hard-drive health data spanning 10 years. We also propose a new statistic - a confidence margin which produces a point estimate, error-margin and confidence interval that gives a range of RUL values during which the hard-drive may fail. In a future study, we intend to train and understand the transformer's performance on other manufacturers. Another potential direction we want to explore is to make the transformer more resilient against cyber attacks. Deep learning models are vulnerable to adversarial attacks such as data poisoning attacks or tampering of weights. Architectural hardening is one way to make these models robust against such adversaries.

\section*{Acknowledgment}
The research reported in this publication was supported by the Division of Research and Innovation at San Jos\'e State University under Award Number 23-UGA-08-044. The content is solely the responsibility of the author(s) and does not necessarily represent the official views of San Jos\'e State University.

\bibliographystyle{IEEEtran} 
\bibliography{main}

\end{document}